# Path Planning under Time-Dependent Uncertainty

Michael P. Wellman, Matthew Ford, and Kenneth Larson
Artificial Intelligence Laboratory
University of Michigan
1101 Beal Avenue
Ann Arbor, MI 48109-2110 USA
{wellman, mford, kenlars}@eecs.umich.edu

## Abstract

Standard algorithms for finding the shortest path in a graph require that the cost of a path be additive in edge costs, and typically assume that costs are deterministic. We consider the problem of uncertain edge costs, with potential probabilistic dependencies among the costs. Although these dependencies violate the standard dynamic-programming decomposition, we identify a weaker stochastic consistency condition that justifies a generalized dynamic-programming approach based on stochastic dominance. We present a revised path-planning algorithm and prove that it produces optimal paths under time-dependent uncertain costs. We test the algorithm by applying it to a model of stochastic bus networks, and present empirical performance results comparing it to some alternatives. Finally, we consider extensions of these concepts to a more general class of problems of heuristic search under uncertainty.

## 1 INTRODUCTION

Most research on state-space search in Artificial Intelligence and Operations Research assumes that the cost of a path is the sum of the cost of its edges, and that the cost of edges are known with certainty. The standard search algorithms (best-first (priority-first) search, A*) rely on these assumptions for their optimality guarantees. Nonlinear path costs, or equivalently, probabilistic and dependent path costs, can defeat these algorithms, as we see below.

In this paper, we focus on the problem of probabilistic and dependent path costs. We demonstrate that given a relatively weak form of stochastic consistency (monotonicity) condition, slightly modified versions of the standard algorithms will produce optimal results. The modification is that instead of storing only the best path to each intermediate node, we now store all *admissible* paths, where admissibility is based on stochastic dominance. Although this can increase complexity significantly, because we prune the dominated paths, the typical search is far short of exhaustive.

In the following section, we define the problem and present an example illustrating the difficulty with using the standard path-planning algorithms. Subsequent sections introduce the concept of stochastic consistency, develop the modified algorithm, and establish its optimality properties. We then illustrate the algorithm by applying it to a stochastic bus-network model, and present some empirical results on randomly generated examples. Finally, we discuss extended applicability of the basic concepts introduced.

## 2 PATH PLANNING UNDER TIME-DEPENDENT UNCERTAINTY

Consider a transportation network with $V$ nodes denoting locations and $E$ edges denoting possible transportation operations between the locations connected. If travel times are static (that is, the duration of a trip from $a$ to $b$ does not depend on departure time), then we can compute the fastest route from any given origin to all possible destinations using Dijkstra's well-known shortest-path algorithm, where the costs on each link are the travel times. This algorithm has a worst-case complexity of $O(V^2)$.[1] The basic idea behind this algorithm is dynamic programming—we start from the origin and systematically find the best paths to all other nodes, using the previously-found best paths. This procedure exploits the *optimality principle*, which dictates that the best path from $a$ to $b$ through $c$ must be comprised of the best paths from $a$ to $c$ and $c$ to $b$.

If the travel times are stochastic but independent (that is, the distribution of travel times for one link does not depend on the actual travel time on others), then the route with the fastest *expected* total travel time can be found similarly with Dijkstra's algorithm, where the costs on each link correspond to expected travel times.

---

[1] For sparse graphs, a simple variant of the algorithm has a running time of $O(E \log V)$.

Unfortunately, this shortest-path algorithm is not universally valid when the travel times are *time-dependent*. This sort of situation should be expected in realistic highway networks, where traffic patterns vary throughout the day, as well as in other transportation networks (e.g., bus routes), where transfer times depend on fixed schedules. Buses and other scheduled services are time dependent because the time to get from $a$ to $b$ depends on which bus we catch and how long we have to wait, which depends on when we arrive at $a$. Time dependence can arise in both the deterministic and stochastic cases.

For an example of stochastic time dependence, suppose we are planning a trip from Ann Arbor, Michigan, to Montreal, Quebec, and intend to take the Canadian portion of the trip by train from Windsor, Ontario. The transportation network for this example is depicted in Figure 1. It is noon, and we have two options for the journey from Ann Arbor to Windsor. First, we could take the bus. The bus is very reliable, and is guaranteed to arrive at the Windsor depot at exactly 13:30. Our second option is to call Leadfoot Taxi Service, which has the potential to get us there earlier, but with some uncertainty. This is because the driver's aggressive highway tactics may reduce the travel time, but we run the risk of getting pulled over by the Detroit Autobahn Patrol, or delayed by suspicious customs agents at the border. Specifically, the taxi's travel time is uniformly distributed between 70 and 120 minutes, that is, our arrival at Windsor is $\sim U[13:10, 14:00]$.

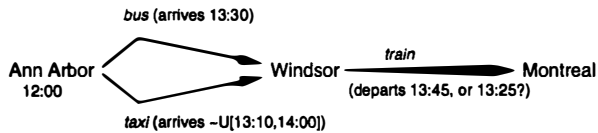

Figure 1: A simple transportation network with time-dependent uncertainty.

Given that our aim is to minimize the expected travel time to Montreal, which option should we take? The expected arrival of Leadfoot at Windsor is 13:35, whereas the expected arrival of the bus is of course 13:30. This suggests that the bus might be preferred. Suppose that the next eastbound train from Windsor departs at 13:45, and the subsequent one is scheduled for several hours later. By taking the bus, we are guaranteed to make the next train, but with Leadfoot, we will miss the train with probability 0.3. In this case, the bus is indeed the better choice. But suppose the next train were scheduled to leave at 13:25 instead. In this event, the bus would be guaranteed to *miss* the connection, whereas Leadfoot would offer us an opportunity to catch the train with probability 0.3. Thus, we see that the best path from Ann Arbor to Windsor depends on the schedule from Windsor to Montreal, and hence the optimality principle is violated. Without the decomposition that would be justified by this principle, it appears that we must in general explore all paths from origin to destination, an exponentially complex prospect.

Let us return briefly to the deterministic case. For deterministic time dependence, it has been shown (Kaufman & Smith 1993) that the standard shortest-path algorithm is indeed sound as long as the network satisfies the following reasonable consistency condition. Let $s$ and $t$ be departure times such that $s \leq t$, and let $c_{ij}(x)$ denote the time-dependent cost (travel time) of traveling from location $i$ to location $j$ at time $x$. The network is *consistent* iff for all $i$ and $j$,

$$s + c_{ij}(s) \leq t + c_{ij}(t). \qquad (1)$$

This consistency condition seems quite reasonable for time-dependent transportation networks. It merely says that although leaving later can perhaps reduce the duration of traversing an edge, it cannot decrease the ultimate arrival time. Given this condition, the principle of optimality underlying Dijkstra's algorithm applies, and the shortest-path problem can be solved relatively efficiently.

A stochastic version of (1), with the times replaced by expectations, would similarly validate the use of the standard algorithm with expectations. That is, if $E[s] \leq E[t]$ implies that for all $i$ and $j$,

$$E[s + c_{ij}(s)] \leq E[t + c_{ij}(t)],$$

then the optimality principle would be preserved. However, this version would not be reasonable, as demonstrated by the example above. There the bus has an earlier expected arrival time at Windsor, but depending on the train schedule may not lead to an earlier expected arrival time at Montreal. Hall (1986) presents some other simple examples where the condition above is violated, and consequently, where the expectation version of the shortest-path algorithm would produce suboptimal paths.

## 3   STOCHASTIC CONSISTENCY

As an alternative, we propose the following condition. Let $c_{ij}(x)$ denote the time-dependent travel time (a random variable) from location $i$ to location $j$ given departure at time $x$. Let us say the network is *stochastically consistent* iff for all $i$, $j$, $s \leq t$, and $z$,

$$\Pr(s + c_{ij}(s) \leq z) \geq \Pr(t + c_{ij}(t) \leq z). \qquad (2)$$

In other words, the probability of arriving by any given time $z$ cannot be increased by leaving later. This appears to be the most natural (and most benign) generalization of the deterministic consistency condition (1). It is clearly satisfied by our example above, as long as the later trains are not expected to pass the earlier ones. And if they were (for example, an express train), we could recover stochastic consistency by simply modifying our policy at Windsor station to wait for the best train. Indeed, as long as waiting at a location is allowed without penalty, we can impose stochastic consistency on any network.[2]

Equation (2) is based on the concept of *stochastic dominance* (Stoyan 1983; Whitmore & Findlay 1978), a common way to extend an ordering relation to random variables. Taking arrival time at a node as a random variable, one arrival distribution dominates another iff its cumulative probability function is uniformly greater than or equal to that of the other. It can be shown (see, for example, Fishburn and Vickson (1978)) that this is equivalent to the statement that the dominant arrival time has a lower expectation than the other, for any monotone transformation of the arrival times.[3] What this means is that the dominant arrival time is preferred regardless of the degree of time preference, whether it is linear in arrival time or even a deadline function (i.e., where all we care about is arrival by a given time), as long as utility is nonincreasing in duration.

This condition justifies a modified version of the shortest-path algorithm, where instead of maintaining the shortest path found to all intermediate nodes (in the uncertain case, a probability distribution of travel times), we maintain all undominated paths. If one path to a node dominates another (in the sense of stochastic ordering), then the stochastic consistency condition ensures that the latter cannot be part of an overall shortest path. This generalized use of the optimum principle can lead to substantial savings if the network contains many dominated paths, as we would expect.

## 4   PRIORITY-FIRST SEARCH WITH DOMINANCE PRUNING

The following algorithm is a generalized variant of priority-first search (PFS), in which we prune search by dominance rather than simple comparison of path costs. To accommodate this generalization, we must maintain a set of undominated paths with each node encountered, rather than a single best found so far. The algorithm, called **PFS-Dominance**, is described by the following elements:

**Input**: a transportation network with uncertain time-dependent edge costs, a nonincreasing utility function $u$, an origin (optionally with a specified starting time), and a destination.

**Output**: a sequence of nodes, corresponding to a path of edges taken from the origin to the destination.

**Data structures**:

- PQ, a priority queue. Items associate a node with a path and a path cost, with path costs prioritized by expectation under $u$.[4]
- Closed list, a lookup table. Associates nodes with the undominated paths found thus far to that node.

**Procedure**:

1. Add origin to PQ with path cost 0.
2. Get highest-priority item from PQ. If this item has lower expected utility than a known path to the destination, then terminate and return the best path (the one maximizing $u$) to the destination found so far.
3. Add item to closed list. If there is already another path to that node with dominating priority, then do nothing and go to step 2. Otherwise, add the path and its cost to the closed-list associated with this node.
4. Generate successors to this item. Construct new paths for each possible bus we could take from this node, and put the resulting items on PQ. Go to step 2.

Note that this algorithm is just like basic PFS, except for the dominance criterion and termination condition.

**Theorem 1** *If the network is stochastically consistent and $u$ is nonincreasing, then PFS-Dominance will find the optimal path from the origin to the destination.*

*Proof*: The algorithm explores the the space of paths exhaustively, except for two classes of paths: (i) paths pruned by stochastic dominance, and (ii) paths not yet explored when the termination condition becomes satisfied. To establish the result, we show that the algorithm selects the best path among those explored, and that the unexplored paths can never be optimal.

Let us consider first the paths pruned by stochastic dominance. Suppose $P_1$ and $P_2$ are two paths to node $i$, and that $P_1$ dominates $P_2$. Let $P_1(x)$ denote the probability of arriving at node $i$ by time $x$ using path $P_1$, and $P_2(x)$ the corresponding probability for path $P_2$. Suppose we extend these paths by adding the edge from $i$ to $j$. The probability that we will arrive at $j$ by time $z$ using the extension of path $P$ is given by

$$\int \Pr(x + c_{ij}(x) \leq z) dP(x), \qquad (3)$$

---
[2]For example, we can simply delete the local train from the network, or, treat the express train as going directly to the distal node (which is of course what it does). It is possible to conceive of situations where waiting is not allowed (e.g., communication networks with limited storage buffers at the nodes), but for most realistic transportation applications we can think of, the condition is quite reasonable.

[3]When dealing with utilities, rather than costs, the directions of relationships in these statements would be reversed.

[4]We could use any nonincreasing function of arrival time as priority (e.g., simple expectation, or earliest possible arrival time) for the exploration part of the algorithm, but using $u$ simplifies the termination condition.

Because the network is stochastically consistent (2), we know that the inner probability expression is a non-increasing function of $x$. And since $P_1$ stochastically dominates $P_2$, this implies that the value of (3) with $P_1$ substituted for $P$ exceeds the corresponding value with $P_2$. Moreover, since this holds for any $z$, we have that the extended version of $P_1$ stochastically dominates that of $P_2$. By repeating this argument, we see that for any extension of path $P_2$, there is a corresponding extension of $P_1$ at least as good. Therefore, $P_2$ can never be the prefix of a uniquely optimal path, and we can prune it safely.

The algorithm terminates when an existing path to the destination has higher expected utility than the path at the top of PQ. Since the network is stochastically consistent, extending paths on PQ can only lower their expected utility. And since the existing path has higher expected utility than the highest priority (expected utility) path on the PQ, it is preferred to all paths on the PQ, as well as all their possible extensions.

When the above condition is met, we are left with a set of undominated paths to the destination. Since we have eliminated only inadmissible alternatives, picking the best one of these yields the optimal path. QED.

The above theorem establishes that stochastic dominance is a valid pruning criterion for stochastically consistent networks. In fact, it can also be shown that for any more aggressive pruning criterion, we could construct a stochastically consistent network in which the discarded path's extension is arbitrarily better than the one retained. In this sense, stochastic dominance is exactly the right pruning relation for this class of problems.[5]

## 5  EMPIRICAL STUDIES

### 5.1  THE BUS NETWORK MODEL

In order to test the PFS-Dominance algorithm, we adopted a simple model of scheduled bus service, based on the model developed by Hall (1986). Consider a directed graph in which a single edge represents a series of buses traveling from one node to another. The travel time of a single bus in this model is assumed to follow the exponential (EXP) distribution, which defined by two parameters, $M$ and $\lambda$. The distribution for $x$, the bus's arrival time, is given by:

$$\text{EXP}[M, \lambda] = \begin{cases} \frac{1}{\lambda} e^{-(x-M)/\lambda} & \text{if } x \geq M \\ 0 & \text{otherwise} \end{cases}$$

We can integrate this function from $M$ to $t$ to obtain the probability that the bus will arrive by time $t$:

$$\Pr(x \leq t) = 1 - e^{-(t-M)/\lambda} \quad (4)$$

From these definitions, we see that $M$ represents the earliest possible arrival time for a bus, and $\lambda$ the average time of arrival beyond $M$. That is, the expected arrival time for a bus is $M + \lambda$. Note that there is no finite time by which the bus is guaranteed to arrive.

Given the arrival model above, a bus is described by a departure time, $d$, along with the $(M, \lambda)$ parameters. Clearly, we must stipulate that $d \leq M$, as the bus cannot arrive at the next stop before it leaves. An edge in the bus network is described as a sequence of buses, $\langle(d_1, M_1, \lambda_1), \ldots, (d_k, M_k, \lambda_k)\rangle$. In order to make this graph stochastically consistent (2), we must ensure that for a particular edge with multiple buses, each bus dominates all subsequent buses.

Let $A$ and $B$ be buses with parameters $(M_A, \lambda_A)$ and $(M_B, \lambda_B)$, respectively. Recall that dominance holds iff the cumulative probability of arrival by $t$ (equation (4)) for $A$ is at least as great as for $B$, for all $t$. It is easy to show that $A$ dominates $B$ iff $M_A \leq M_B$ and $\lambda_A \leq \lambda_B$ (see (Stoyan 1983)). The former condition is perhaps obviously necessary; the earliest arrival time must be earlier for the dominating bus. The latter condition is required because (4) is strictly decreasing in $\lambda$, and in fact if $\lambda_B < \lambda_A$ then there is some $t$ for which bus $B$ has higher cumulative arrival probability, regardless of the $M$ values.

We can therefore ensure stochastic consistency for the bus network by requiring for each edge, that $d_i < d_j$ implies $M_i \leq M_j$ and $\lambda_i \leq \lambda_j$. Given this condition, it is always optimal to take the next departing bus on a given edge, regardless of the time of arrival at the bus stop.

One final issue is what to do if we arrive at a node after all the buses have left. This is always possible, due to the unboundedness of the exponential distribution. In our model, we assume that the traveler can choose to walk, which would take time $W \sim \text{EXP}[\bar{M}, \bar{\lambda}]$. We further assume that $\bar{M}$ is high enough so that one would never choose the walking option over a bus, and that $\bar{\lambda}$ is sufficiently great to assure stochastic consistency.

### 5.2  PATH COSTS

To apply the priority-first search method, we need a way to calculate the distribution of travel times for *sequences* of buses. First, let us consider a two-edge path, from the origin to node $a$, and then from node $a$ to node $b$. Because the departure time from the origin is fixed, we simply take the next bus departing for $a$. However, we cannot determine which of the buses from $a$ to $b$ will be taken, because the arrival time at $a$ is uncertain. Therefore, in order to calculate the arrival time at $b$, we consider each bus we could take from $a$, weighted by the probability that we will take it. These weights are simply the probabilities that we will arrive at $a$ during the intervals in which each bus is next. Therefore the path cost to $b$ can be described by a list of exponentials, associated with scalar probabilities. Because there is some finite probability that we will miss all of the buses, we must also include this remainder probability of incurring the walking cost $W$.

---

[5]This does not definitively establish the optimality of the pruning algorithm, because it is based on only a pairwise comparison. By examining the set of admissible paths (e.g., pruning paths dominated by *convex combinations* of existing paths), we could perhaps generate fewer paths without sacrificing optimality.

Now consider an extension of this path to node $c$. We can extend the path cost in a similar manner, using the path cost to $b$ to calculate the probability of taking each bus, and associating that probability with the parameters of the respective bus to $c$. The result (for any number of edges) is a list of the same form—a sequence of triples $(p, M, \lambda)$, describing the ultimate bus in the path along with the probability of taking it.

We have noted above the criterion for stochastic dominance of individual buses. But to apply the algorithm PFS-Dominance we need to evaluate stochastic dominance of path costs, which are weighted sequences of buses. We illustrate this concept with an example. Let $C_1$ and $C_2$ be two path costs.

$$\begin{aligned} C_1 &= \langle(0.5, 5, 5), (0.3, 15, 5)\rangle, \\ C_2 &= \langle(0.44, 10, 5), (0.26, 20, 5)\rangle. \end{aligned}$$

That is, $C_1$ is distributed EXP[5, 5] with probability 0.5, and EXP[15, 5] with probability 0.3. It has walking cost $W$ with the remaining probability 0.2.

Although $C_2$ includes a bus that dominates one in $C_1$, and vice versa, it is nevertheless the case that $C_1$ stochastically dominates $C_2$. That is, for any time $z$, the probability that one arrives at the destination by time $z$ is greater under path cost $C_1$ than under $C_2$. This relationship becomes clearer if we lay out the two path costs on a number line. In Figure 2, $C_1$ and $C_2$ are arranged on top of each other, sorted by bus departure time, with the width of each bus proportional to its probability.

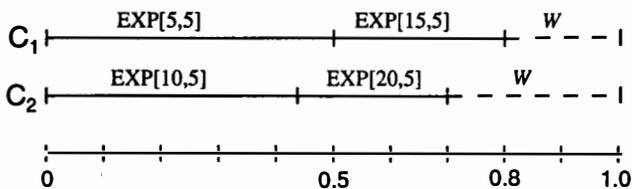

Figure 2: Stochastic dominance of distributions of exponentials.

To determine dominance of the overall costs, we test dominance of each vertical slice. Because this network is stochastically consistent, ordering the buses by departure time is tantamount to ordering them by dominance. Therefore, to evaluate dominance of $C_1$ over $C_2$, we can simply evaluate individual bus dominance at each probability threshold of $C_1$.

Given these techniques and definitions, we now have all the components of PFS-Dominance for the bus-network model. We can extend the model slightly by also allowing constant edge costs, to represent walking between a pair of nodes.

## 5.3 COMPUTATIONAL RESULTS

To test the PFS-Dominance algorithm on this bus model, we randomly generated a series of networks of varying size. Each network is an $n \times n$ grid, with each edge assigned a sequence of buses defined by the three parameters $(m, \lambda, f)$: minimum trip time, average delay, and frequency. Frequency dictates the departure times for the various buses, and a bus departing at time $d$ (some multiple of $f$) is distributed $EXP[d + m, \lambda]$. The grid structure we generate is illustrated in Figure 3. In our tests, the origin is at the lower left, and the destination at the upper right.

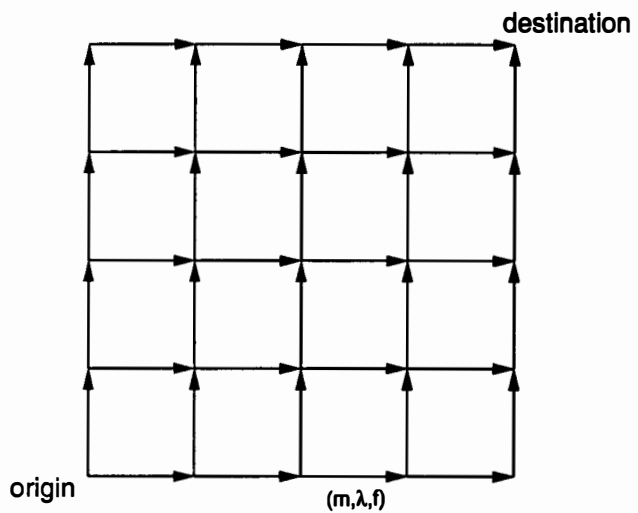

Figure 3: A grid of buses.

To evaluate the algorithm, we counted the number of paths explored for grids of different sizes. As a baseline, we note that the total number of paths from origin to destination in an $n \times n$ grid is[6]

$$\binom{2n-2}{n-1}.$$

In Figure 4, we plot the path count for square grids of sizes up to 17. Each point represents one randomly generated network. We also attempted to run the PFS algorithm without dominance pruning, but were unable to solve problems of size greater than $n = 8$, due to space limitations. Note that we counted all paths generated, including intermediate ones, which explains how the non-dominance algorithm can generate more paths than the total number to the destination.

Thus we see that dominance pruning permits us to solve problems of considerably greater size than would be possible with exhaustive search. Of course, pruning by expectation (i.e., using the standard shortest-path algorithm) will generate a much smaller (polynomial) number of paths, but can produce suboptimal results. In general, we can trade off path cost for computational efficiency by adopting more strident pruning criteria (i.e., weakening the dominance criterion).

---

[6]To see this, observe that every path has length exactly $2n - 2$, defined by how we choose the $n - 1$ points at which we go "up".

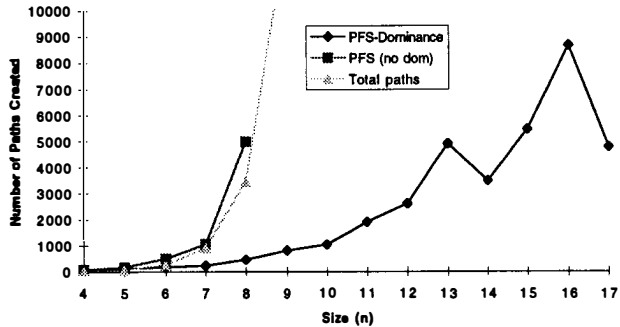

Figure 4: The number of paths generated by PFS-Dominance for grids of varying size. The nonmonotonicity of this measure with $n$ is likely attributable to our small sample size (one randomly generated network for each value).

## 6 ADAPTIVE PATH PLANNING

In the version of the stochastic bus problem presented above, the traveler constructs a complete path in advance, and executes it in an open-loop manner. However, in many realistic situations, the traveler may make decisions along the way about which bus to take, based on the uncertainty resolved along the way. In particular, the traveler typically knows what time it is when arriving at a bus stop, and this may influence the optimal edge to take from that node.

In closed-loop, or *adaptive* path planning, the planning algorithm produces a policy as output, rather than a single path. For any arrival time at any node, the policy specifies which outgoing edge to take. In other words, the product is a conditional plan. This plan should have a higher expected utility than the optimal open-loop plan, in general, but may be more complex to construct.

To compare the quality of conditional and unconditional plans, we implemented an adaptive path planning algorithm based on dynamic programming, and tested it on our bus-network model. The algorithm, **Adaptive-Path**, is described below.

**Input**: an acyclic transportation network, a nonincreasing utility function $u$, and a destination.[7]

**Output**: a policy for each node, consisting of a partitioning of the set of arrival times from 0 to infinity, where each interval in the partition is associated with an outgoing edge.

**Procedure**:

1. Let all nodes have no policy, except the destination node, and nodes without outgoing edges, which are assigned degenerate policies.
2. If all nodes have policies, terminate and return the policies for all nodes.
3. Choose a node with no policy, whose successors all have policies.
4. Partition the set of arrival times from 0 to infinity into a finite set of intervals, where each interval is started by a possible departure, and ends at the next departure opportunity.[8]
5. For each interval, calculate the expected utility of each departure option (e.g., each bus that departs after the interval), and select the option with maximum expected utility. Because we have already computed policies for all of the successor nodes, calculating these expected utilities is straightforward. Associate the optimal choice and its expected utility with the interval.
6. Go to step 2.

Because it is simply dynamic programming applied backward from the destination, this algorithm produces optimal policies as long as the network is acyclic. We applied this algorithm to a $10 \times 10$ bus network, using a deadline utility function. In the deadline model, utility is unity if we arrive before the deadline, and zero if we arrive afterward. Thus, expected utility is simply the probability of making the deadline. We ran the algorithm for deadlines between 150 and 250 minutes, at one-minute intervals. The results are plotted in Figure 5.

For comparison with the open-loop case, we ran the algorithms **PFS-Dominance** and the standard shortest-path algorithm (PFS with pruning based on expected value) on the same 101 problems.[9] Deadlines near 150 are nearly impossible to meet for any algorithm, and deadlines greater than 230 are very easy. In intermediate cases, we notice differences among the algorithms. As expected, the adaptive algorithm always performs best, and the expected-value algorithm the worst. **PFS-Dominance** outperforms the latter because it produces the optimal path, but this falls short of the optimal policy.

---

[7]The origin and starting time are not inputs because the algorithm computes optimal policies for all possible starting points.

[8]Here we exploit the fact that our buses leave nodes at fixed scheduled times. Some constraint of this sort is required in order to keep the policies finite.

[9]Actually, only **Adaptive-Path** required 101 runs. For **PFS-Dominance** we ran the search once, terminating when the top of PQ was stochastically dominated by an existing path. Then we used the various deadlines to select from the same admissible set. For the expected-value case, only one run was required. The deadline criterion is ignored and we use the same path every time.

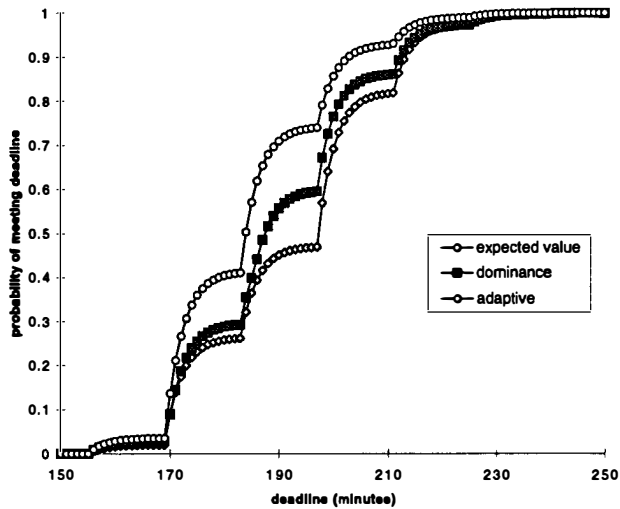

Figure 5: Probabilities of meeting deadlines for three path-planning algorithms. Kinks in the performance curves correspond to the earliest arrival times of buses; when a new bus becomes feasible the probability of making a deadline increases substantially.

We have also considered partially adaptive path planning, where uncertainty is resolved at some nodes but not others. The result is a hybrid algorithm, which uses the adaptive algorithm where allowed, and **PFS-Dominance** elsewhere. In its open-loop planning, **PFS-Dominance** treats any adaptive node as a destination. If the network is acyclic, this hybrid algorithm produces the optimal policy using the available information. As expected, with partial adaptivity the results are intermediate in value between the pure open- and closed-loop cases.

## 7 RELATED WORK

Although there has been a great deal of previous work on deterministic versions (static and dynamic) of vehicle routing problems, relatively little effort (until recently) has been devoted to stochastic formulations of these tasks. This is true especially in the case of problems with multiple objectives (e.g., travel time, safety, expense, route familiarity), which have rarely been analyzed under stochastic assumptions (see (List et al. 1991) for the few exceptions).

Polychronopoulos (1992) presents a taxonomy of stochastic shortest-path problems, in which the main distinction is what the uncertainty is about:

1. which edges are available to traverse,
2. which vertex results from traversing an edge, or
3. the edge costs.

For example, Qi and Poole (1991) address the first kind of uncertainty, as does much work in theoretical computer science. Our work addresses the last type, which we believe to be the most prevalent form of uncertainty in path-planning problems. Indeed, determinism in state transitions might be regarded as *necessary* for regarding our problem as *path* planning; otherwise we are not really choosing a path at all.

A number of works have focused on simplifying distributions over edge costs, so that Dijkstra's algorithm can then be applied. However this generally works for limited classes of utility functions, and produces arbitrarily suboptimal results for others. For example, Kamburowski (1985) discusses reducing the distribution of a path to an "optimality index" using measures such as mean, variance, the probability that the path takes the minimum possible time, and the variance of the path from the minimum possible time. Eiger, et al. (1985) show how utility functions that are linear or exponential in edge costs validate the optimality principle, and thus Dijkstra's algorithm.

The approach taken here is a version of *generalized dynamic programming* (Carraway, Morin, & Moskowitz 1990), which is essentially dynamic programming with a totally ordered criterion replaced with a partial order. We have cast generalized DP as PFS with dominance-pruning, which leads to a concise algorithm description. The partial-order concept has been most often been applied to multicriterial optimization. Stewart and White extend this to include a heuristic estimate of distance to a goal, in their multiobjective version of A* (Stewart & White 1991).

Likewise, Loui (1983) uses dominance to solve multi-attribute, deterministic shortest path problems. Mirchandani and Soroush (1985) use mean/variance dominance to solve the problem when the cost distribution of arcs can be uniquely described by the mean and variance (this possibility had also been noted by Loui).

Given all this prior work, the specific contribution of this paper is a formulation of the stochastic consistency condition, and identifying stochastic dominance as the appropriate pruning criterion to apply in the **PFS-Dominance** algorithm. The idea of stochastic consistency is related to our previous work on qualitative probabilistic influences (Wellman 1990); the condition can be interpreted as saying that taking any edge is a positive influence on arrival time.

The idea of imposing a consistency condition was inspired by the work of Kaufman and Smith (1993) (noted above) in the deterministic context. Bard and Bennett (1991) use an approximation of stochastic dominance to reduce the search in stochastic networks; however, they apply the dominance condition in a preprocessing stage, rather than incrementally during search, and do not exploit the stochastic consistency condition. Moreover, their heuristic use of "near dominance" can be made to yield arbitrarily suboptimal paths, depending on the utility function.

Finally, as mentioned above, our bus-network model is based on the work of Hall (1986). Hall notes the diffi-culty of the time-dependent, stochastic shortest path problem, and proposes an algorithm that generates all paths in order of earliest possible arrival, terminating when this earliest-possible measure exceeds the expected time of a path already found. We suspect that this algorithm is far less efficient on average than **PFS-Dominance**, but it appears that this algorithm has never been implemented, and we have not yet ourselves carried out a systematic comparison.

## 8 DISCUSSION AND EXTENSIONS

Although our exposition has focused on particular transportation optimization problems, what we are really interested in is extending state-space search methods to dramatically broader problem classes. The assumption of additive costs is typically not seriously examined by AI authors, probably due to the focus on deterministic problems, and because it was unclear how to relax the assumption anyway. In extending standard AI techniques to the uncertain case, we seek to do so in as broad a manner as possible. Thus, handling special cases such as Gaussian distributions or completely independent edge costs is of only limited interest. In this work, we have identified a qualitative, widely applicable, monotonicity condition (stochastic consistency) that in turn justifies a generic pruning method (stochastic dominance), each of which can be employed regardless of the analytic form or representational formalism used to express uncertainty.

In ongoing work, we intend to extend these concepts to other search frameworks. For example, with a heuristic estimate of remaining distance, the **PFS-Dominance** algorithm becomes a stochastic version of A* (analogous to multiobjective A* (Stewart & White 1991)). Indeed, we have implemented such an algorithm, and have tested it on a stochastic version of the 8-puzzle, where the costs of moving tiles is uncertain and the utility function nonlinear. Results from this work will be reported at a later date.

We would also like to explore the possibility that some search problems formulated with uncertain state transitions could be reformulated as path planning problems (i.e., deterministic state transitions), with the uncertainty relegated to edge cost. If such a transition would preserve the stochastic consistency condition without unduly blowing up the state space, our methods could prove advantageous. Right now, we have only speculation.

Finally, we also contemplate further refinements, such as developing space-limited versions of the algorithm using techniques from recent AI work.

### Acknowledgements

This work was supported in part by Grant F49620-94-1-0027 from the Air Force Office of Scientific Research, and an NSF National Young Investigator award. We would also like to thank Robert Smith, Izak Duenyas, Chip White, and Runping Qi for helpful exchanges.

## References

Bard, J. F., and Bennett, J. E. 1991. Arc reduction and path preference in stochastic acyclic networks. *Management Science* 37:198–215.

Carraway, R. L.; Morin, T. L.; and Moskowitz, H. 1990. Generalized dynamic programming for multicriteria optimization. *European Journal of Operational Research* 44:95–104.

Eiger, A.; Mirchandani, P. B.; and Soroush, H. 1985. Path preferences and optimal paths in probabilistic networks. *Transportation Science* 19:75–84.

Fishburn, P. C., and Vickson, R. G. 1978. Theoretical foundations of stochastic dominance. In Whitmore and Findlay (1978).

Hall, R. W. 1986. The fastest path through a network with random time-dependent travel times. *Transportation Science* 20(3):182–188.

Kamburowski, J. 1985. A note on the stochastic shortest route problem. *Operations Research* 22:696–698.

Kaufman, D. E., and Smith, R. L. 1993. Fastest paths in time-dependent networks for intelligent vehicle-highway systems application. *IVHS Journal* 1(1).

List, G. F.; Mirchandani, P. B.; Turnquist, M. A.; and Zografos, K. G. 1991. Modeling and analysis for hazardous materials transportation: Risk analysis, routing/scheduling, and facility location. *Transportation Science* 25:100–114.

Loui, R. P. 1983. Optimal paths in graphs with stochastic or multidimensional weights. *Communications of the ACM* 26:670–676.

Mirchandani, P. B., and Soroush, H. 1985. Optimal paths in probabilistic networks: A case with temporary preferences. *Transportation Science* 12:365–381.

Polychronopoulos, G. H. 1992. *Stochastic and Dynammic Shortest Distance Problems*. Ph.D. Dissertation, Massachusetts Institute of Technology. Technical Report 199.

Qi, R., and Poole, D. 1991. High level path planning with uncertainty. In *Proceedings of the Seventh Conference on Uncertainty in Artificial Intelligence*, 287–294.

Stewart, B. S., and White, III, C. C. 1991. Multiobjective A*. *Journal of the ACM* 38:775–814.

Stoyan, D. 1983. *Comparison Methods for Queues and Other Stochastic Models*. John Wiley and Sons.

Wellman, M. P. 1990. Fundamental concepts of qualitative probabilistic networks. *Artificial Intelligence* 44:257–303.

Whitmore, G. A., and Findlay, M. C., eds. 1978. *Stochastic Dominance: An Approach to Decision Making Under Risk*. Lexington, MA: D. C. Heath and Company.